%% file: anonymous-submission-latex-2025.tex
\documentclass[letterpaper]{article} 
\usepackage{aaai25}  
\usepackage{times}  
\usepackage{helvet}  
\usepackage{courier}  
\usepackage[hyphens]{url}  
\usepackage{graphicx} 
\usepackage{natbib}  
\usepackage{caption} 
\usepackage{algorithm}

\usepackage{newfloat}
\usepackage{listings}

\usepackage{latexsym}

\usepackage[T1]{fontenc}

\usepackage[utf8]{inputenc}

\usepackage{microtype}

\usepackage{multirow}
\usepackage{appendix}
\usepackage{lipsum,booktabs}
\usepackage{color}
\usepackage{subcaption}

\usepackage{algpseudocode}

\usepackage{booktabs}
\usepackage{multirow}

\input{math_commands.tex}

\DeclareCaptionStyle{ruled}{labelfont=normalfont,labelsep=colon,strut=off} 
\floatstyle{ruled}
\newfloat{listing}{tb}{lst}{}
\floatname{listing}{Listing}
\title{C3oT: Generating Shorter Chain-of-Thought without Compromising Effectiveness}
\author{
    Yu Kang, 
    Xianghui Sun, 
    Liangyu Chen \footnote{The corresponding author.}, 
    Wei Zou
}
\affiliations{
    Beike Inc., Beijing, China\\
    \{kangyu009, sunxianghui002, chenliangyu003, zouwei026\}@ke.com
}

\usepackage{bibentry}

\begin{document}

\maketitle

\begin{abstract}
\input{abstract}
\end{abstract}

%

\vspace{-0.3cm}
\section{Introduction}
\label{sec:intro}

\input{intro}

\vspace{-0.2cm}
\section{Related Work}
\label{sec:related}
\input{related}

\vspace{-0.3cm}
\section{Method}
\label{sec:method}
\vspace{-0.1cm}
\input{method}

\section{Experiment}
\label{sec:experiment}
\input{experiment}

\vspace{-0.2cm}
\section{Analysis}
\label{sec:analysis}
\input{analysis}

\section{Conclusion}
\label{sec:conclusion}
\input{conclusion}

\bibliography{aaai25}

\clearpage
\appendix

\section{Appendix}
\label{sec:appendix}
\input{appendix}

\end{document}

%% file: math_commands.tex
\usepackage{amsmath,amsfonts,bm}

\def\eqref#1{equation~\ref{#1}}

\def\1{\bm{1}}

\DeclareMathAlphabet{\mathsfit}{\encodingdefault}{\sfdefault}{m}{sl}
\SetMathAlphabet{\mathsfit}{bold}{\encodingdefault}{\sfdefault}{bx}{n}



%% file: abstract.tex

Generating Chain-of-Thought (CoT) before deriving the answer can effectively improve the reasoning capabilities of large language models (LLMs) and significantly improve the accuracy of the generated answer. However, in most cases, the length of the generated CoT is much longer than the desired final answer, which results in additional decoding costs. Furthermore, existing research has discovered that shortening the reasoning steps in CoT, even while preserving the key information, diminishes LLMs' abilities. These phenomena make it difficult to use LLMs and CoT in many real-world applications that only require the final answer and are sensitive to latency, such as search and recommendation. To reduce the costs of model decoding and shorten the length of the generated CoT, this paper presents \textbf{C}onditioned \textbf{C}ompressed \textbf{C}hain-of-\textbf{T}hought (C3oT), a CoT compression framework that involves a compressor to compress an original longer CoT into a shorter CoT while maintaining key information and interpretability, a conditioned training method to train LLMs with both longer CoT and shorter CoT simultaneously to learn the corresponding relationships between them, and a conditioned inference method to gain the reasoning ability learned from longer CoT by generating shorter CoT. We conduct experiments over four datasets from arithmetic and commonsense scenarios, showing that the proposed method is capable of compressing the length of generated CoT by up to more than 50\% without compromising its effectiveness.

%% file: intro.tex
The Chain-of-Thought (CoT) \cite{nye2021show,marasovic2021few,CoTPrompting,stepbystep,lampinen2022can} methodology significantly augments the reasoning abilities of large language models (LLMs), providing critical capabilities for sub-task decomposition in complex problem-solving scenarios. 
Furthermore, models trained with rich signals, including reasoning processes; explanation traces; and step-by-step thought processes, generally exhibit superior performance \cite{orca,mitra2023orca}. 
While answering after thinking can elicit highly effective generations by activating LLMs' reasoning abilities, the intermediate reasoning steps in model outputs are often much longer than the desired final answers, notably increasing the cost during the inference phase, and hindering the model's employment in many real-world applications, such as search and recommendation, which usually focus only on the final answer and is sensitive to latency. Therefore, striking a balance between the demand for fast decoding in LLMs applications and the need for long reasoning steps has become an urgent issue.

However, recent studies indicate that lengthening the reasoning steps in CoT considerably enhances LLMs' reasoning abilities across multiple tasks. Alternatively, shortening the reasoning steps, even while preserving the key information, significantly diminishes the reasoning abilities of models \cite{jin2024impact}. \citet{fuyaoComplexity} propose a complexity-based method for CoT selection and find that CoT with higher reasoning complexity, i.e., chains with more reasoning steps, achieve substantially better performance on multi-step reasoning tasks. Similar conclusions have been drawn from \citet{expresssivePower}'s work, which explores the relationship between the capabilities of LLMs and the number of CoTs' reasoning steps, varying from logarithmic, linear, to polynomial, based on input length. They also found that the increase in LLMs' computational power depends crucially on the amount of intermediate reasoning steps added.

There has been little work \cite{deng2023implicit,liu2024expediting} focused on compressing the length of generated CoT without sacrificing model performance. Implicit-CoT \cite{deng2023implicit} attempted to use LLMs' internal hidden states to perform implicit reasoning, replacing explicitly producing the CoT reasoning steps, but the results of this method is still significantly falling behind the explicit CoT method.

Based on these results, we ask: \textit{Is there a method that can significantly reduce the length of intermediate reasoning steps in generated CoT without compromising effectiveness?} The answer is \textit{yes}, we propose \textbf{C}onditioned \textbf{C}ompressed \textbf{C}hain-of-\textbf{T}hought (C3oT), a CoT compression framework, to achieve this goal. Specifically, we first present a CoT compressor to condense the original complex CoT into their shortest form while retaining essential information and interpretability, now we have pairs of longer CoT and shorter CoT. 
We further introduce a conditioned training method to train LLMs with both longer CoT and shorter CoT simultaneously, and by conditioning longer CoT and shorter CoT using distinct initial prompt tokens before instructions, LLMs can learn the differences and connections between them.
Lastly, we propose the conditioned inference method which is used in the inference phase, by applying the initial prompt tokens used for conditioning the shorter CoT before instructions, LLMs can generate CoT with significantly shorter length during inference while maintaining the accuracy of the derived final answer.

To validate the effectiveness of our approach, we conduct experiments on four datasets from two domains that require reasoning, i.e., arithmetic (GSM8K, MathQA) and commonsense (ECQA, StrategyQA). The results show that our method's performance is on par with models trained using only the original longer CoT across all datasets, while significantly shortening the length of generated CoT. Additionally, we design extensive experiments and discussions to analyze the contribution of different components in our approach, as well as to explore future research directions of CoT compression based on our method.

The contributions of this paper are as follows:
\begin{itemize}
    \item We propose C3oT, a CoT compression framework used to reduce the cost of model inference by drastically shortening the length of CoT generated by LLMs without loss of effectiveness. We are the first to significantly shorten the length of CoT in model outputs without sacrificing performance, filling the gap in the field of model inference acceleration in terms of shortening the length of intermediate output.
    \item Comprehensive experiments demonstrate that our CoT compression method outperforms all baselines on various reasoning tasks, such as math reasoning (GSM8K, MathQA) and commonsense reasoning (ECQA, StrategyQA). We conduct detailed ablation studies and analyses to prove the effectiveness of our approach.
    \item We conduct a series of extension experiments based on the proposed C3oT framework, providing insights for future research directions in the field of CoT compression and further demonstrating the effectiveness of our method.
\end{itemize}

%% file: related.tex
\subsection{LLMs Inference Acceleration}
Due to the conflict between the outstanding performance of LLMs and the difficulty of their application in real-world scenarios, an increasing amount of research is focusing on accelerating the inference of LLMs.
These works primarily focus on reducing the number of input tokens to be processed in order to reduce the inference cost. Some approaches focus on reducing the length of prompts by using prompt tuning to learn special tokens \cite{wingate2022prompt,chevalier2023adapting,ge2023context,mu2024learning}. Some approaches attempt to compress prompt based on information entropy \cite{li2023compressing,llmlingua,jiang2023longllmlingua} and data distillation \cite{pan2024llmlingua}. Some studies utilize LLMs to summarize input dialog or data, transforming them into efficient memory and knowledge \cite{langchain,zhang2023mlcopilot}. There are also some studies focus on token pruning or token merging \cite{kim2022learned,modarressi2022adapler,bolya2022token}.

However, during the inference stage, the cost of decoding and generating output by the model is significantly higher than the cost of processing the input. Therefore, as the enhancements of model capabilities brought by CoT gain increasing attention, the additional cost involved in generating CoT cannot be ignored. Nevertheless, accelerating the generation of CoT has not received widespread attention.

Recently, only Implicit-CoT \cite{deng2023implicit} has attempted to accelerate the generation of CoT. It uses the hidden states of different layers in LLMs to perform implicit reasoning based on knowledge distillation, avoiding the explicit generation of CoT and thereby accelerating the inference process. But Implicit-CoT severely sacrifices model performance. The results generated by this method significantly fall behind those of the explicit CoT method.

The method proposed in this paper also aims to accelerate the inference by reducing the length of generated CoT. By utilizing conditioned training method, the model is enabled to learn both longer and shorter CoT simultaneously, and during the conditioned inference phase, the model is able to stimulate the reasoning capabilities learned from the longer CoT by generating a shorter CoT. In this way, our approach significantly reduces the length of generated CoT without compromising the effectiveness of the model.

It's worth mentioning that there is a line of studies that attempt to accelerate inference through model quantization \cite{dettmers2022gpt3,frantar2022optq,xiao2023smoothquant}, pruning \cite{frantar2023sparsegpt,sun2023simple,das2023beyond}, and other similar techniques. These methods are orthogonal to ours and can be used together.

\vspace{-0.2cm}
\subsection{Chain-of-Thought Analyzing}
There is some research focusing on exploring the relationship between the length of CoT and its effects. Interestingly, all of these studies \cite{fuyaoComplexity,expresssivePower,jin2024impact} have found that lengthening the intermediate reasoning steps in the CoT can enhance LLMs' capabilities. \citet{fuyaoComplexity} find that CoT with higher reasoning complexity, i.e., chains with more reasoning steps, achieve substantially better performance on multi-step reasoning tasks. \citet{expresssivePower} explore the relationship between the capabilities of LLMs and the number of CoTs' reasoning steps, varying from logarithmic, linear, to polynomial, based on input length, and they find that the increase in LLMs' computational power depends crucially on the amount of intermediate reasoning steps added. \citet{jin2024impact} design experiments that expand and compress the reasoning steps in CoT while keeping all other factors constant. They find that lengthening the reasoning steps even without adding new information considerably enhances LLMs' reasoning abilities. Conversely, shortening the reasoning steps while preserving the key information, significantly diminishes the reasoning abilities of models.

This paper also focuses on the relationship between the length of CoT and its effectiveness, but we propose a method that can significantly compress the length of generated CoT without compromising its effectiveness.

%% file: method.tex
\subsection{Problem Statement}
Given a dataset $\{(x_{i}, r_{i}^{long}, y_{i})\}_{i=1}^{N}$, where $x_{i}$ denotes the instruction, $y_{i}$ is its corresponding answer and $r_{i}^{long}$ is a well-designed detailed CoT for deriving the answer. We consider a compressor $\mathcal{F}$ that systematically compress any input CoT to its shortest form $r^{short} = \mathcal{F}(r^{long})$, retaining only the key information. Our goal is to train an LLM on $\mathcal{D} = \{(x_{i}, r_{i}^{long}, r_{i}^{short}, y_{i})\}_{i=1}^{N}$ so that during inference, the distribution of generated answers derived from compressed, shorter CoT is as similar to answers derived from original, longer CoT as possible.

\subsection{Conditioned Compressed Chain-of-Thought (C3oT)}
\label{subsec:c3ot}
\input{fig_1_main_fig}
Next, we elaborate on the details of the C3oT framework, which shortens the generated CoT during inference without compromising its effectiveness. An overview of the proposed framework is shown in Figure~\ref{fig:main_fig}.

\noindent \textbf{Compressor} \indent The CoT compressor can be any summarization model that processes the input text to only retain its core information and returns a condensed version. In this paper, we employ GPT-4 \cite{achiam2023gpt} as the compressor for CoT compression. 

Using GPT-4 as the compressor is also to validate the conclusions of previous research. Specifically, even when using the current most powerful closed-source model to ensure that the compressed CoT retains all key information and interpretability while merely removing redundant words, only using these compressed, shorter CoT to derive the answers will still affect the model's performance. This is consistent with the conclusions of previous research and further proves the value of our approach.

We prompt GPT-4 to obtain the corresponding compressed, shorter CoT $r_{i}^{short}$ for all original, longer CoT $r_{i}^{long}$ in the dataset $\{(x_{i}, r_{i}^{long}, y_{i})\}_{i=1}^{N}$, and compose $\mathcal{D} = \{(x_{i}, r_{i}^{long}, r_{i}^{short}, y_{i})\}_{i=1}^{N}$. We also investigate the impact on our approach of employing different models as compressors in the Analysis section. The Appendix contains all the prompts we use.

\noindent \textbf{Conditioned Training} \indent Inspired by OpenChat \cite{wang2023openchat}, we can regard $\mathcal{D}$ as a class-conditioned dataset $\mathcal{D}_{c} = \{(x_{i}, \tilde{r}_{i}, y_{i}, c_{i})\}$. Each instruction $x_{i}$ in the dataset corresponds to both a longer CoT and a shorter CoT. The CoT of different lengths are distinguished through different conditions:
\begingroup\makeatletter\def\f@size{10}\check@mathfonts
\def\maketag@@@#1{\hbox{\m@th\large\normalfont#1}}%
\begin{align*}
\tilde{r}_{i} = \begin{cases}
r_{i}^{long}  & \text{ if } c_{i}=long \\
r_{i}^{short}  & \text{ if } c_{i}=short
\end{cases}
\end{align*}
\endgroup

\noindent where $\tilde{r}_{i}$ can be either $r_{i}^{long}$ or $r_{i}^{short}$, controlled by condition $c_{i}$.

Then we fine-tune an LLM on $\mathcal{D}_{c}$ to teach it the relationship between $r_{i}^{long}$ and $r_{i}^{short}$. We model the LLM to be fine-tuned as a class-conditioned policy $\pi_{\theta}(y, \tilde{r} \vert x, c)$. This can be easily implemented by conditioning each CoT which belongs to either a longer CoT or a shorter CoT using distinct initial prompt tokens before instruction as shown below:
\begingroup\makeatletter\def\f@size{10}\check@mathfonts
\def\maketag@@@#1{\hbox{\m@th\large\normalfont#1}}%
\begin{align*}
\begin{matrix}
\textbf{[Long Condition]}  &
\begin{matrix}
\text{Answer and provide a}
\\
\text{detailed thought process:}
\end{matrix}
\\
\\
\textbf{[Short Condition]}  &
\begin{matrix}
\text{Answer and provide as brief a}
\\
\text{thought process as possible:}
\end{matrix}
\end{matrix}
\end{align*}
\endgroup

After adding conditions in this way, each sample in the original dataset, for example:

\vspace{0.2cm}

\noindent \textit{\textbf{Instruction:} Natalia sold clips ... How many clips did Natalia sell altogether in April and May?}

\noindent \textit{\textbf{Rationale:} ... Natalia sold 48+24 = <<48+24=72>>72 clips altogether in April and May.}

\vspace{0.2cm}

\noindent becomes two samples containing the original, longer CoT and the compressed, shorter CoT, respectively:

\vspace{0.2cm}

\noindent \textit{\textbf{Instruction:} Answer and provide a detailed thought process: Natalia sold clips ... How many clips did Natalia sell altogether in April and May?}

\noindent \textit{\textbf{Rationale:} ... Natalia sold 48+24 = <<48+24=72>>72 clips altogether in April and May.}

\vspace{0.2cm}

\noindent and

\vspace{0.2cm}

\noindent \textit{\textbf{Instruction:} Answer and provide as brief a thought process as possible: Natalia sold clips ... How many clips did Natalia sell altogether in April and May?}

\noindent \textit{\textbf{Rationale:} ... She sold 72 clips in April and May.}

\vspace{0.2cm}

Now, fine-tuning an LLM on $\mathcal{D}_{c}$ is the same as general supervised fine-tuning (SFT). It is worth mentioning that during fine-tuning, longer CoT and shorter CoT in $\mathcal{D}_{c}$ do not need to appear in pairs, and there is no need to use any method to inform the LLM how they correspond to each other, just randomly shuffle $\mathcal{D}_{c}$ and train the model like general SFT. Additionally, our conditioned training method can also be regarded as a data augmentation method, but it does not introduce any extra knowledge \cite{maini2024rephrasing}.

\noindent \textbf{Conditioned Inference} \indent During the inference phase, we assume that the model after conditioned training has learned the differences and connections between the longer CoT and the shorter CoT. Considering that we aim to apply the fine-tuned model to a real-world application and exclusively generate shorter CoT to derive the needed final answer efficiently, we use the same specific prompts that were employed in shorter CoT during the conditioned training phase as below:
\begingroup\makeatletter\def\f@size{10}\check@mathfonts
\def\maketag@@@#1{\hbox{\m@th\large\normalfont#1}}%
\begin{align*}
\begin{matrix}
\textbf{[Inference Prompt]}  &
\begin{matrix}
\text{Answer and provide as brief a thou-}
\\
\text{ght process as possible: <Question>}
\end{matrix}
\end{matrix}
\end{align*}
\endgroup


%% file: fig_1_main_fig.tex
\begin{figure*}[!hbt]
    \vspace{-0.2cm}
    \begin{center}
    \includegraphics[width=\textwidth]{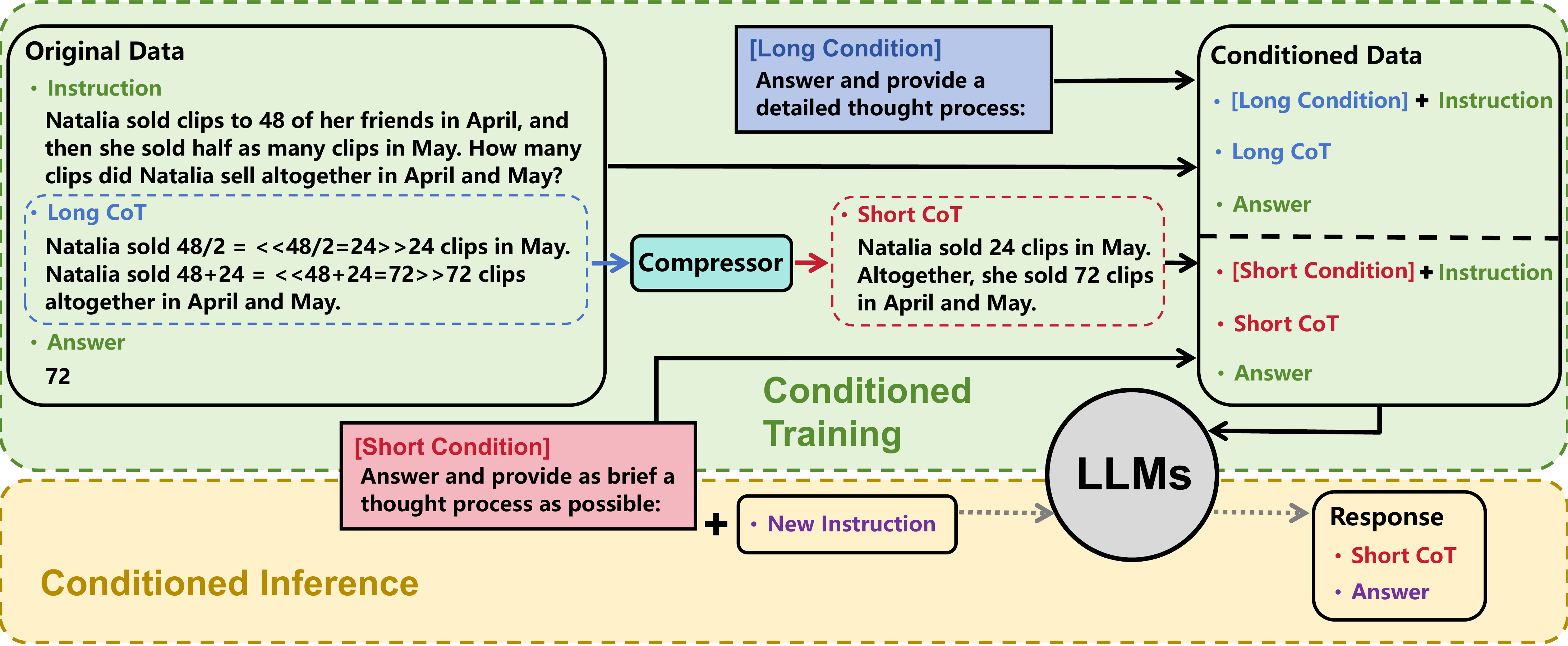}
    \end{center}

    \vspace{-0.3cm}

    \caption{The overall framework of C3oT. Solid arrows denote the conditioned training phase, while dashed arrows denote the conditioned inference phase.}
    \label{fig:main_fig}
    \vspace{-0.5cm}
\end{figure*}

%% file: experiment.tex
\subsection{Settings}
\noindent \textbf{Datasets} \indent To comprehensively validate the effectiveness of C3oT, we evaluated its performance across four datasets from two domains. For math reasoning, we use \textbf{GSM8K} \cite{cobbe2021training} and \textbf{MathQA} \cite{amini2019mathqa}. As for commonsense reasoning, we use \textbf{ECQA} \cite{aggarwal2021explanations} and \textbf{StrategyQA} \cite{geva2021did}. All these datasets not only contain the final answers but also include the carefully human-designed CoT used to arrive at the final answers. We followed the training and testing set division as outlined in the original paper of the dataset used, trained C3oT on the training set, and evaluated its performance on the test set, excluding StrategyQA. Due to the inaccessibility of ground truths for the StrategyQA test set, we proceeded to further split the original StrategyQA training set into training and test sets. Please refer to Appendix for more details.

\noindent \textbf{Implementation Details} \indent In this paper, we train C3oT based on \texttt{LLaMA-2-Chat-7B} and \texttt{-13B} \cite{touvron2023llama}. We fine-tune the model for 2 epochs on each dataset using the AdamW optimizer with a sequence length of 2,048 tokens and a batch size of 128. The AdamW optimizer's hyperparameters are set as follows: $\beta_{1}=0.9,\beta_{2}=0.999,\epsilon=10^{-6}$, and weight decay of 0.001. We employ a cosine learning rate schedule with a maximum learning rate of $1 \times 10^{-5}$.

\noindent \textbf{Baselines} \indent We consider the following baselines:
\begin{itemize}
    \item \textit{Short CoT}: Use GPT-4 as the compressor to compress the original, longer CoT as much as possible while retaining key information and interpretability, then train models using only these compressed, shorter CoT. All the prompts we employed are displayed in the Appendix.
    \item \textit{Long CoT}: Use only the original, longer CoT to train models.
    \item \textit{Implicit-CoT} \cite{deng2023implicit}\footnote{For a fair comparison, we do not use the data synthesis methods in the original paper.}: Use the LLM's internal hidden states to perform implicit reasoning, instead of explicitly producing the CoT reasoning steps. The implicit reasoning steps are distilled from a teacher model trained on explicit CoT reasoning.
\end{itemize}

\noindent All the baselines are also trained based on \texttt{LLaMA-2-Chat} \texttt{-7B} and \texttt{-13B}, including the teacher model and student model of Implicit-CoT. The hyperparameters used for training baselines are the same as those in training C3oT.

\noindent \textbf{Evaluation} \indent We measure the following metrics:
\begin{itemize}
    \item \textit{Accuracy}: Following previous works \cite{BBH,cobbe2021training,amini2019mathqa,aggarwal2021explanations,geva2021did}, we measure accuracy using exact match, computed by comparing the generated final answer with the ground-truth label. 
    \item \textit{Compression Rate}: Additionally, we measure the compression rate to evaluate the reduction in length of the generated CoT. The compression rate $\rho$ is defined as $\rho = (L - \tilde{L}) / L$, $\rho \in \left ( - \infty ,1 \right ]$, where $L$ is the length of the CoT generated by \textit{Long CoT}, which we regard as the benchmark length. And $\tilde{L}$ is the length of the generated compressed CoT. A larger value of compression rate implies a greater reduction in length, resulting in a lower inference cost, which is preferable. When no intermediate CoT steps are generated, the compression rate can reach $1$. If the generated CoT is even longer than $L$, the compression rate becomes negative.
\end{itemize}

\subsection{Main Results}
\input{table_1_main_res}
Table~\ref{tab:main_res} reports the results of our approach alongside baselines on GSM8K, MathQA, ECQA and StrategyQA. It can be seen that our proposed C3oT consistently outperforms the Implicit-CoT in accuracy by a large margin in all experiments. Implicit-CoT successfully avoids explicitly generating CoT, achieving a 100\% compression rate. However, it significantly sacrifices the model's performance, which is not what we want. In addition, three other conclusions can be drawn from these results:

Firstly, comparing the results of \textit{Short CoT} and \textit{Long CoT} reveals that even while using the most powerful model GPT-4 as the compressor to preserve the key information and interpretability in the compressed, shorter CoT in the training sets, 
it still significantly diminishes the model's effectiveness. This conclusion is consistent with previous studies.

Secondly, while shortening the length of generated CoT reduces the model's performance across all datasets, the degree of performance decrease varies across different datasets. Tasks requiring more reasoning abilities, such as math, experience a greater decrease in performance, whereas tasks with lower reasoning demands, such as commonsense, see a relatively smaller decrease. Similarly, although \textit{C3oT} has achieved similar or even better performance than \textit{Long CoT} on all datasets, there is still a slight lag in mathematical tasks, while in commonsense tasks, it even surpasses or significantly outperforms \textit{Long CoT}. This is still related to the reasoning ability required for the tasks.

Lastly, the compression rates on four datasets show that when preserving the key information and interpretability in the compressed, shorter CoT in the training sets, and keeping the compressor unchanged, the compression rate is only related to the dataset itself and not significantly influenced by the domain of the task. In other words, if the original, longer CoT in the training set is more detailed, containing more redundant information, then the compression rate achievable by \textit{C3oT} will be higher. Additionally, we further analyze the impact of different compressors on compression rates in the Analysis section.

%% file: table_1_main_res.tex
\begin{table*}[!hbt]

    \vspace{-0.2cm}

    \small
    \centering

    \begin{tabular}{@{}cccccccccc@{}}
    \toprule
    \multirow{3}{*}{\begin{tabular}[c]{@{}c@{}}Model\\ Size\end{tabular}} & \multicolumn{1}{c|}{\multirow{3}{*}{Method}} & \multicolumn{4}{c}{Arithmetic}                                                                                                                                                                      & \multicolumn{4}{c}{Commonsense}                                                                                                                                                \\ \cmidrule(l){3-10} 
                                                                          & \multicolumn{1}{c|}{}                        & \multicolumn{2}{c}{GSM8K}                                                                        & \multicolumn{2}{c|}{MathQA}                                                                      & \multicolumn{2}{c}{ECQA}                                                                         & \multicolumn{2}{c}{StrategyQA}                                              \\ \cmidrule(l){3-10} 
                                                                          & \multicolumn{1}{c|}{}                        & Acc            & \multicolumn{1}{c|}{\begin{tabular}[c]{@{}c@{}}Compression\\ Rate\end{tabular}} & Acc            & \multicolumn{1}{c|}{\begin{tabular}[c]{@{}c@{}}Compression\\ Rate\end{tabular}} & Acc            & \multicolumn{1}{c|}{\begin{tabular}[c]{@{}c@{}}Compression\\ Rate\end{tabular}} & Acc            & \begin{tabular}[c]{@{}c@{}}Compression\\ Rate\end{tabular} \\ \midrule
    \multirow{4}{*}{7B}                                                   & Short CoT                                    & 31.01          & 58.63                                                                           & 46.16          & 29.53                                                                           & 61.93          & 53.41                                                                           & 67.59          & 37.99                                                      \\
                                                                          & Long CoT                                     & \textbf{37.38} & 0                                                                               & \textbf{51.46} & 0                                                                               & 63.96          & 0                                                                               & 69.66          & 0                                                          \\
                                                                          & Implicit-CoT                                 & 11.16          & 100                                                                             & 14.62          & 100                                                                             & 21.14          & 100                                                                             & 30.01          & 100                                                        \\
                                                                          & C3oT                                         & \textbf{36.92} & 56.67                                                                           & 50.35          & 27.39                                                                           & \textbf{69.38} & 51.55                                                                           & \textbf{72.41} & 42.04                                                      \\ \midrule
    \multirow{4}{*}{13B}                                                  & Short CoT                                    & 42.46          & 59.52                                                                           & 52.97          & 29.85                                                                           & 66.79          & 55.27                                                                           & 74.83          & 47.79                                                      \\
                                                                          & Long CoT                                     & \textbf{48.07} & 0                                                                               & \textbf{56.21} & 0                                                                               & 68.92          & 0                                                                               & \textbf{76.21} & 0                                                          \\
                                                                          & Implicit-CoT                                 & 14.36          & 100                                                                             & 17.00          & 100                                                                             & 23.54          & 100                                                                             & 35.77          & 100                                                        \\
                                                                          & C3oT                                         & \textbf{47.10} & 57.78                                                                           & \textbf{56.62} & 31.04                                                                           & \textbf{71.93} & 55.28                                                                           & \textbf{76.55} & 44.56                                                      \\ \bottomrule
    \end{tabular}

\vspace{-0.2cm}

\caption{The Accuracy (\%) and Compression Rate (\%) performance of the proposed C3oT and baselines. The \textbf{bold} scores denote the best performance, as well as performances within 1\% of the best.}

\label{tab:main_res}

\vspace{-0.3cm}

\end{table*}

%% file: analysis.tex
In this section, we conduct experiments to answer the following questions, in order to analyze the contributions of different components in our approach, and further explore more future research directions for CoT compression based on the proposed C3oT framework.

\subsection{What is the contribution of the class-conditioned policy?}
\input{table_2_conditions}
We conduct an ablation study on the conditions of C3oT to ascertain the contribution of the class-conditioned policy. For \textit{w/o condition}, we remove the distinct initial prompt tokens before instruction 
and treat the data containing longer CoT and shorter CoT as equivalent, and then fine-tune the models in the regular supervised fine-tuning (SFT) manner.

Table~\ref{tab:conditions} shows that \textit{C3oT} outperforms \textit{w/o condition} in terms of both accuracy and compression rate across all datasets. This is because without the class-conditioned policy, language models lack explicit signals to discern between the longer CoT and the shorter CoT, and during training, the models not only fail to learn the differences and connections between the longer CoT and the shorter CoT, 
but also get confused by two kinds of CoT with significantly different lengths, thus affecting the performance.

\subsection{What is the impact of different compressors on C3oT?}
\label{subsec:compressors}
\input{table_3_analysis}
To investigate the impact of different compressors on our method, we conduct experiments comparing the results of using GPT-4 as the compressor with the results of using the open-source models \texttt{LLaMA-2-Chat-7B} and \texttt{-13B} as the compressors.
To distinguish the results of different compressors, in Table~\ref{tab:analysis}, we name the results as \textit{Short CoT$_{<Compressor>}$} and \textit{C3oT$_{<Compressor>}$}. It is worth mentioning that \textit{Short CoT$_{GPT-4}$} and \textit{C3oT$_{GPT-4}$} are precisely \textit{Short CoT} and \textit{C3oT} in Table~\ref{tab:main_res}.
The prompts used for the different compressors are the same.

Comparing the results of \textit{Short CoT$_{GPT-4}$} and \textit{Short CoT$_{LLaMA2\_\ast}$} as well as \textit{C3oT$_{GPT-4}$} and \textit{C3oT$_{LLaMA2\_\ast}$} in Table~\ref{tab:analysis}, it's evident that the compressed, shorter CoT generated by LLaMA-2 in the training set are slightly inferior in quality to those generated by GPT-4, resulting in a minor decrease in model accuracy, though not significantly. However, the conciseness of the compressed, shorter CoT generated by LLaMA-2 in the training set are noticeably poorer than those generated by GPT-4, leading to a compression rate that is over 15\% lower. This indicates that preserving the key information in the original, longer CoT is not difficult for the compressor, but the more powerful the compressor, the more it can produce concise compressed CoT.

We display several compressed CoT in the training set corresponding to different compressors in the Appendix.

\subsection{Is C3oT effective for expanded CoT?}
Previous studies have shown that lengthening the reasoning steps in CoT can enhance the model's reasoning abilities. Therefore, in this part, we explore whether our approach can compress the much longer, expanded CoT and maintain its effectiveness. We follow the 5 reasoning steps expansion methods (\textbf{Think About The Word}, \textbf{Read the Question Again}, \textbf{Repeat State}, \textbf{Self-Verification} and \textbf{Make Equation}) proposed by \citet{jin2024impact} and use GPT-4 to expand the original CoT. The much longer, expanded CoT is then combined with the compressed, shorter CoT generated by GPT-4 that we used above to form a class-conditioned dataset for training C3oT, which we name \textit{C3oT$_{Expansion}$}. In parallel, we refer to the results obtained from the model trained using only the much longer, expanded CoT as \textit{Expanded CoT}.
The expansion prompt and examples of expanded CoT in the training set can be found in the Appendix.

Comparing the results of \textit{Long CoT} and \textit{Expanded CoT} in Table~\ref{tab:analysis}, it is evident that lengthening the reasoning steps in CoT does improve the model's reasoning abilities, which is consistent with previous studies. While comparing the results of \textit{C3oT$_{GPT-4}$} and \textit{C3oT$_{Expansion}$}, we observe that although the improvement is not as significant as from \textit{Long CoT} to \textit{Expanded CoT}, \textit{C3oT$_{Expansion}$} still manages to achieve a better result than \textit{C3oT$_{GPT-4}$} while maintaining a similar compression rate. This not only demonstrates the effectiveness of our approach on the much longer, expanded CoT but also presents a method to enhance the model's reasoning abilities without incurring additional costs.

\subsection{What is the impact of training sets with different compression rates on C3oT?}
\label{subsec:compression_rates}
\input{fig_2_multi_compress_rate}
When we use GPT-4 as the compressor in the previous sections, we prompt it to retain as much key information and interpretability from the original CoT in the training set as possible, the compression rate of the compressed CoT in the resulting GSM8K training set is about 50\% compared to the original CoT (the detailed metrics can be found in the Appendix). We wonder about the impact on model performance when all restrictions on the compressor are removed, and it is only required to compress the original CoT in the training set to a specified compression rate. Specifically, we compress the original CoT in the GSM8K training set with compression rates varying from 50\% to 100\% (no CoT) in 10\% increments.

The results are shown in Figure~\ref{fig:multi_compress_rate_7b} and Figure~\ref{fig:multi_compress_rate_13b}. Firstly, we observe that the accuracy of \textit{C3oT} decreases as the compression rate of the training set increases, indicating that \textit{C3oT} is not effective across all compression levels. Secondly, although \textit{C3oT} outperforms \textit{Short CoT} at almost all training set compression rates, the performance gap between the two widens as the compression rate decreases, only to narrow again at the 50\%. 

These two phenomena indicate that \textit{C3oT} can only achieve results comparable to \textit{Long CoT} if the compressed, shorter CoT in the training set retains sufficient key information, and if the CoT in the training set is over-compressed, using \textit{C3oT} will still lead to a decline in performance. Furthermore, at high compression rates, the compressed, shorter CoT even loses its grammatical structure, rendering it completely uninterpretable. At this point, the results of \textit{C3oT} and \textit{Short CoT} are not significantly different. As the compression rate of the training set decreases, the information and interpretability contained in the compressed CoT increase. Gradually, \textit{C3oT} can leverage the shorter CoT to activate the reasoning abilities learned during the conditioned training phase from longer CoT, thereby widening the performance gap with \textit{Short CoT}. This continues until the information and interpretability in the compressed, shorter CoT reach a certain threshold, satisfying the requirements for \textit{C3oT} to fully activate the CoT's capabilities, achieving results close to those of \textit{Long CoT}. At the same time, the performance of \textit{Short CoT} also sees a significant improvement.

We display several compressed CoT corresponding to different compression rates in the Appendix.

\subsection{What is the impact of mixed conditions training on C3oT?}
In the previous part, we explored the impact of different compression rates of the training set on C3oT. Specifically, during the training phase, for shorter CoT corresponding to various compression rates, we combined them respectively with the longer CoT to form the class-conditioned dataset. Then, we employed the conditioned training method mentioned in the Method section. However, an intuitive idea is to use various distinct initial prompt tokens before instructions to condition shorter CoT corresponding to different compression rates in the training set, and combine them together with the longer CoT to form a mixed class-conditioned dataset.

To investigate this, we implement \textit{Mixed Conditions}. Specifically, we expand the conditions into the following form:
\vspace{-0.05cm}
\begingroup\makeatletter\def\f@size{10}\check@mathfonts
\def\maketag@@@#1{\hbox{\m@th\large\normalfont#1}}%
\begin{align*}
\begin{matrix}
\textbf{[Long Condition]}  &
\begin{matrix}
\text{Answer and provide a}
\\
\text{detailed thought process:}
\end{matrix}
\\
\\
\textbf{[Short Cond.1]}  &
\begin{matrix}
\text{Answer and provide a thought}
\\
\text{process in compression level of 1:}
\end{matrix}
\\
\textbf{.   .   .}
\\
\textbf{[Short Cond.6]}  &
\begin{matrix}
\text{Answer and provide a thought}
\\
\text{process in compression level of 6:}
\end{matrix}
\end{matrix}
\end{align*}
\endgroup

\noindent where Short Cond.1 to Cond.6 denote compression rates of the training set ranging from 50\% to 100\%, respectively.

From Figure~\ref{fig:multi_compress_rate_7b} and Figure~\ref{fig:multi_compress_rate_13b}, we can see that \textit{Mixed Conditions} outperforms \textit{C3oT} across all training set compression rates and even surpasses \textit{Long CoT} at 50\% compression rate. This demonstrates that training with a mix of data at various compression levels through the class-conditioned policy can lead to mutually beneficial effects. 

Moreover, \textit{Mixed Conditions} can generate CoT with different compression levels in the conditioned inference phase by using different initial prompt tokens before instruction. It is worth mentioning that in the Figure~\ref{fig:multi_compress_rate_7b} and Figure~\ref{fig:multi_compress_rate_13b}, the horizontal axis represents the compression rate of the training set for \textit{C3oT}, and for \textit{Mixed Conditions}, it represents the distinct initial prompt tokens before instruction corresponding to that compression rate used during the conditioned inference.

However, just like \textit{C3oT}, \textit{Mixed Conditions} also requires a CoT that contains sufficient information and interpretability to fully leverage the model's reasoning capabilities. When the compression rate is too high, the performance still suffers.

\subsection{Can C3oT select the appropriate compression rate on its own?}
Through the exploration of the previous two parts, we've observed that the length of CoT required to correctly answer questions varies with the difficulty of the questions. For simple questions, the model can provide answers directly even without the CoT (Compression Rate = 100\%). However, as the questions become more challenging, the model needs to undergo a more complex intermediate reasoning process to arrive at the correct answer. Therefore, the overall accuracy of the model tends to decrease as the compression rate increases. In this final part, we aim to explore the capability of C3oT to automatically select the most appropriate compression rate. For the most appropriate compression rate, we refer to the highest CoT compression rate at which the model still can accurately answer questions.

Inspired by Orca 2 \cite{mitra2023orca}, firstly, we arrange the training sets obtained in previous parts by their compression rates, from highest to lowest. Then, we process each training set in order, randomly dividing each into five parts. We use compressed, shorter CoT in four parts to train a model and predict outcomes on the remaining part. This step is repeated three times with different random seeds to obtain three prediction results for each sample in the training set. For each sample, if it is correctly predicted in at least one out of these three attempts, we consider the sample as correctly answerable by the model under the current compression rate and include it in the final training set. Conversely, if a sample is too difficult to be correctly predicted under the current compression rate, it is carried forward to the next round at a lower compression rate for further assessment. Finally, we train C3oT using the conditioned training method mentioned in the Method section with the final training set composed in the above steps and name the result \textit{C3oT$_{Adapt}$} in Table~\ref{tab:analysis}.

The results from Table~\ref{tab:analysis} show that \textit{C3oT$_{Adapt}$} significantly outperforms \textit{C3oT$_{GPT-4}$} in both accuracy and compression rate. This demonstrates that after training C3oT with data at the most appropriate compression rate, the model has learned to autonomously determine the most efficient length of the CoT for questions of varying complexity during the inference phase. This conclusion also opens up a new avenue for the further application of C3oT.

%% file: table_2_conditions.tex
\begin{table*}[!hbt]

    \vspace{-0.2cm}

    \small
    \centering

    \begin{tabular}{@{}cccccccccc@{}}
    \toprule
    \multirow{3}{*}{\begin{tabular}[c]{@{}c@{}}Model\\ Size\end{tabular}} & \multicolumn{1}{c|}{\multirow{3}{*}{Method}} & \multicolumn{4}{c}{Arithmetic}                                                                                                                                                                      & \multicolumn{4}{c}{Commonsense}                                                                                                                                                \\ \cmidrule(l){3-10} 
                                                                          & \multicolumn{1}{c|}{}                        & \multicolumn{2}{c}{GSM8K}                                                                        & \multicolumn{2}{c|}{MathQA}                                                                      & \multicolumn{2}{c}{ECQA}                                                                         & \multicolumn{2}{c}{StrategyQA}                                              \\ \cmidrule(l){3-10} 
                                                                          & \multicolumn{1}{c|}{}                        & Acc            & \multicolumn{1}{c|}{\begin{tabular}[c]{@{}c@{}}Compression\\ Rate\end{tabular}} & Acc            & \multicolumn{1}{c|}{\begin{tabular}[c]{@{}c@{}}Compression\\ Rate\end{tabular}} & Acc            & \multicolumn{1}{c|}{\begin{tabular}[c]{@{}c@{}}Compression\\ Rate\end{tabular}} & Acc            & \begin{tabular}[c]{@{}c@{}}Compression\\ Rate\end{tabular} \\ \midrule
    \multirow{2}{*}{7B}                                                   & C3oT                                         & \textbf{36.92} & \textbf{56.67}                                                                  & \textbf{50.35} & \textbf{27.39}                                                                  & \textbf{69.38} & \textbf{51.55}                                                                  & \textbf{72.41} & \textbf{42.04}                                             \\
                                                                          & w/o condition                                & 34.50          & 35.59                                                                           & 48.07          & 15.86                                                                           & 66.93          & 20.88                                                                           & 70.00          & 9.51                                                       \\ \midrule
    \multirow{2}{*}{13B}                                                  & C3oT                                         & \textbf{47.10} & \textbf{57.78}                                                                  & \textbf{56.62} & \textbf{31.04}                                                                  & \textbf{71.93} & \textbf{55.28}                                                                  & \textbf{76.55} & \textbf{44.56}                                             \\
                                                                          & w/o condition                                & 44.50          & 40.51                                                                           & 55.34          & 18.53                                                                           & 70.54          & 27.19                                                                           & 73.10          & 20.37                                                      \\ \bottomrule
    \end{tabular}

\vspace{-0.2cm}

\caption{Ablation study of the class-conditioned policy (condition) to C3oT.}

\label{tab:conditions}

\vspace{-0.2cm}

\end{table*}

%% file: table_3_analysis.tex
\begin{table}[hbt]

    \small
    \centering

    \begin{tabular}{@{}cccc@{}}
    \toprule
    \multirow{2}{*}{\begin{tabular}[c]{@{}c@{}}Model\\ Size\end{tabular}} & \multicolumn{1}{c|}{\multirow{2}{*}{Method}}                & \multicolumn{2}{c}{GSM8K}                                          \\ \cmidrule(l){3-4} 
                                                                          & \multicolumn{1}{c|}{}                                       & Acc   & \begin{tabular}[c]{@{}c@{}}Compression\\ Rate\end{tabular} \\ \midrule
    \multirow{8}{*}{7B}                                                   & Long CoT                                                    & 37.38 & 0                                                          \\
                                                                          & \begin{tabular}[c]{@{}c@{}}Short CoT$_{GPT-4}$\end{tabular} & 31.01 & 58.63                                                      \\
                                                                          & C3oT$_{GPT-4}$                                             & 36.92 & 56.67                                                      \\ \cmidrule(lr){2-2}
                                                                          & \begin{tabular}[c]{@{}c@{}}Short CoT$_{LLaMA2\_7B}$\end{tabular}  & 31.54 & 42.28                                                      \\
                                                                          & C3oT$_{LLaMA2\_7B}$                                              & 36.13 & 40.82                                                      \\ \cmidrule(lr){2-2}
                                                                          & \begin{tabular}[c]{@{}c@{}}Expanded CoT\end{tabular}   & 39.12 & -310.17                                                    \\
                                                                          & C3oT$_{Expansion}$                                         & 37.30 & 59.27                                                      \\ \cmidrule(lr){2-2}
                                                                          & C3oT$_{Adapt}$                                             & 40.85 & 70.80                                                      \\ \midrule
    \multirow{8}{*}{13B}                                                  & Long CoT                                                    & 48.07 & 0                                                          \\
                                                                          & \begin{tabular}[c]{@{}c@{}}Short CoT$_{GPT-4}$\end{tabular} & 42.46 & 59.52                                                      \\
                                                                          & C3oT$_{GPT-4}$                                             & 47.1  & 57.78                                                      \\ \cmidrule(lr){2-2}
                                                                          & \begin{tabular}[c]{@{}c@{}}Short CoT$_{LLaMA2\_13B}$\end{tabular}  & 40.71 & 40.34                                                      \\
                                                                          & C3oT$_{LLaMA2\_13B}$                                              & 46.53 & 42.48                                                      \\ \cmidrule(lr){2-2}
                                                                          & \begin{tabular}[c]{@{}c@{}}Expanded CoT\end{tabular}   & 49.66 & -307.88                                                    \\
                                                                          & C3oT$_{Expansion}$                                         & 48.12 & 57.66                                                      \\ \cmidrule(lr){2-2}
                                                                          & C3oT$_{Adapt}$                                             & 51.09 & 77.97                                                      \\ \bottomrule
    \end{tabular}

\vspace{-0.2cm}

\caption{Performance of some experiments based on C3oT on GSM8K. The negative compression rates represent the degree of increase in length.}

\label{tab:analysis}

\vspace{-0.3cm}

\end{table}

%% file: fig_2_multi_compress_rate.tex
\begin{figure}
    \begin{subfigure}{0.495\linewidth}
      \centering
      \includegraphics[width=0.98\linewidth]{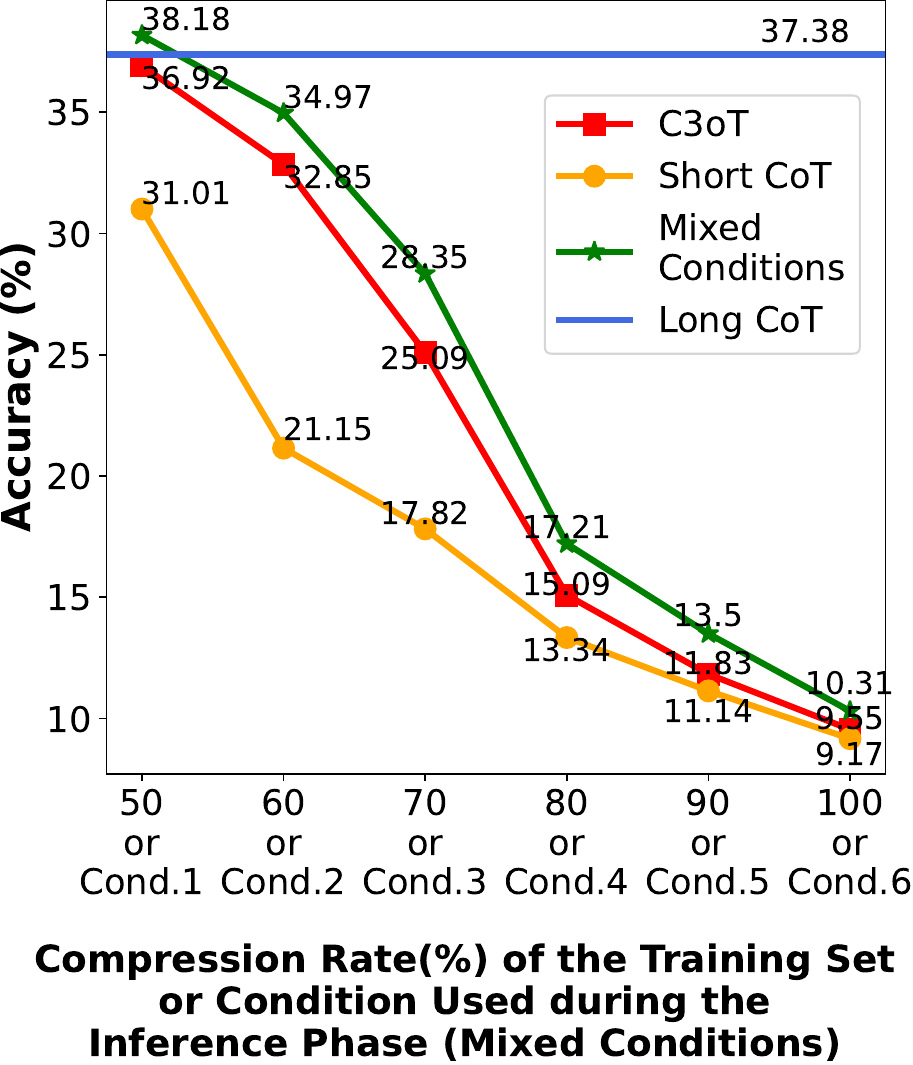}
      \caption{Model Size: 7B}
      \label{fig:multi_compress_rate_7b}
    \end{subfigure}
    \begin{subfigure}{0.495\linewidth}
      \centering
      \includegraphics[width=0.98\linewidth]{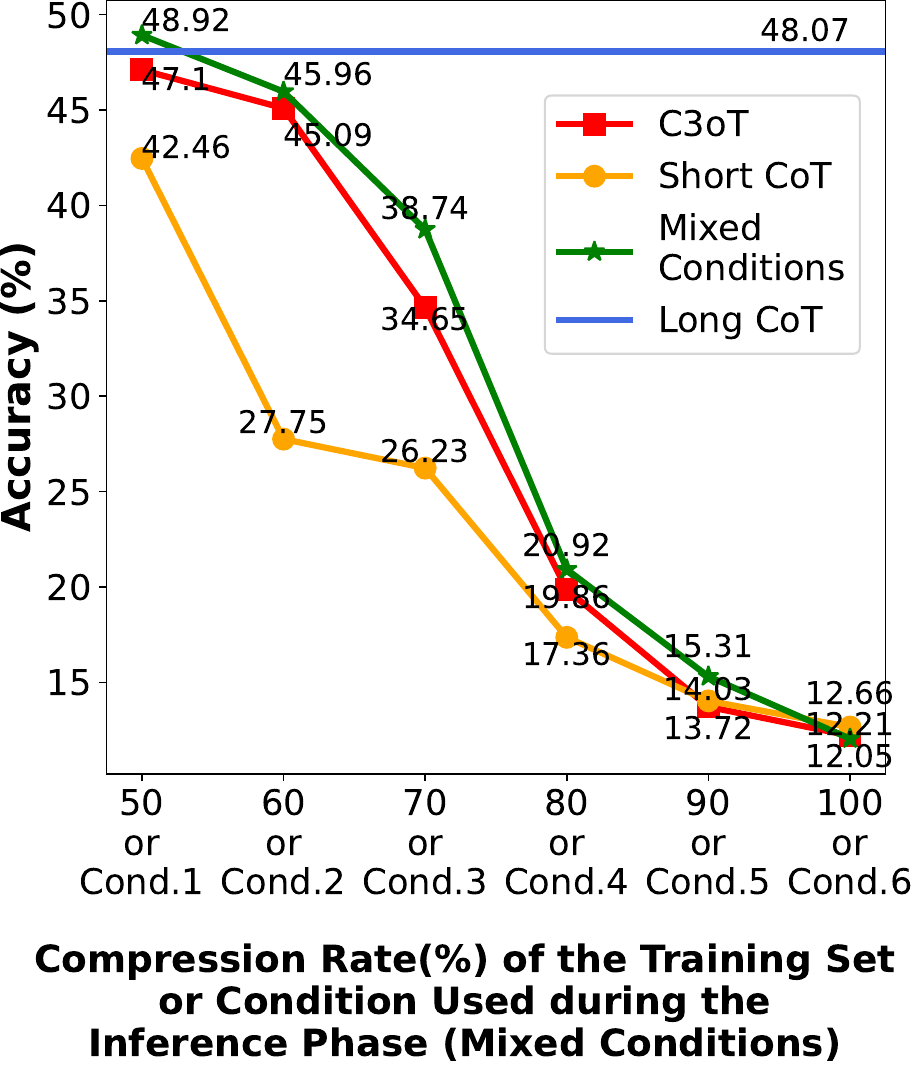}
      \caption{Model Size: 13B}
      \label{fig:multi_compress_rate_13b}
    \end{subfigure}

    \vspace{-0.2cm}

    \caption{Accuracy of different methods vs compression rate of the training set or condition used during the inference phase on GSM8K.}
    \label{fig:multi_compress_rate}
\vspace{-0.5cm}
\end{figure}

%% file: conclusion.tex
We introduce a simple but effective method for CoT compression, named \textbf{C}onditioned \textbf{C}ompressed \textbf{C}hain-of-\textbf{T}hought (C3oT), which is based on the class-conditioned policy. Our approach consists of three modules: Compressor, Conditioned Training, and Conditioned Inference. We validate the effectiveness of our approach on four datasets from two domains that require reasoning, i.e., arithmetic (GSM8K, MathQA) and commonsense (ECQA, StrategyQA), demonstrating that our method's performance is on par with models trained using original longer CoT across all datasets, while significantly shortening the length of generated CoT. Moreover, we conduct experiments to analyze the contribution of different components in our approach. We also observe that more variant approaches to CoT compression can be derived based on C3oT framework, further demonstrating the effectiveness of our method.
Our approach holds significant practical implications, as it enables models, which are trained using complex, longer CoT to enhance reasoning capabilities, to be applied in time-sensitive real-world applications.

%% file: appendix.tex
\subsection{Prompt Templates}
\label{subsec:prompt_templates}
\input{table_append_prompts}

\clearpage

\subsection{Dataset Details}
\label{subsec:dataset_details}
\noindent \textbf{GSM8K} \indent A widely used math reasoning dataset consisting of 8,792 problems. It includes a training set with 7,473 problems and a test set with 1,319 problems, designed to evaluate a model's ability in arithmetic reasoning and formulating mathematical steps using language \cite{cobbe2021training}. In the training set, the average number of tokens in the original CoT is 124. After using GPT-4 as a compressor, the average number of tokens in the compressed CoT is reduced to 56, resulting in a compression rate of 55\% for the training set CoT.

\noindent \textbf{MathQA} \indent A large-scale mathematical reasoning dataset derived from AQuA \cite{ling2017program} through careful selection and processing. It includes a training set with 29,837 problems and a test set with 2,985 problems \cite{amini2019mathqa}. In the training set, the average number of tokens in the original CoT is 91. After using GPT-4 as a compressor, the average number of tokens in the compressed CoT is reduced to 63, resulting in a compression rate of 31\% for the training set CoT.

\noindent \textbf{ECQA} \indent A commonsense reasoning dataset released to advance research in commonsense question answering tasks. It includes a training set with 7,598 problems and a test set with 2,194 problems \cite{aggarwal2021explanations}. In the training set, the average number of tokens in the original CoT is 93. After using GPT-4 as a compressor, the average number of tokens in the compressed CoT is reduced to 39, resulting in a compression rate of 58\% for the training set CoT.

\noindent \textbf{StrategyQA} \indent A multi-hop commonsense reasoning dataset where the required reasoning steps are implicit in the question and should be inferred using a strategy. It includes a training set with 2,290 problems and a private test set with 490 problems. The private test set does not provide ground truth answers or the intermediate reasoning steps from which the answers are derived \cite{geva2021did}. Therefore, to evaluate the model's performance on this dataset, we further split the original training set into a training set with 2,000 problems and a test set with 290 problems. In the training set, the average number of tokens in the original CoT is 54. After using GPT-4 as a compressor, the average number of tokens in the compressed CoT is reduced to 27, resulting in a compression rate of 50\% for the training set CoT.

\clearpage

\subsection{Cases Study}

\label{subsubsec:cases_of_different_compressors}
\input{fig_case_study_compressor}


\label{subsubsec:cases_of_cot_expansion}
\input{fig_case_study_expansion}


\label{subsubsec:cases_of_different_compression_rates}
\input{fig_case_study_rate}

%% file: table_append_prompts.tex
\begin{table}[h]

    \centering

    \begin{tabular}{@{}l@{}}
    \toprule
    \textbf{Prompt For Compression:}                                                                                                                                                                                                                                                                                                                                                                                                                                                                                                    \\ \midrule
    \begin{tabular}[c]{@{}l@{}}You have a question now:\\ \\ QUESTION:\\ \textless{}Here is Instruction\textgreater\\ \\ THOUGHT PROCESS:\\ \textless{}Here is Original CoT\textgreater\\ \\ ANSWER:\\ \textless{}Here is Final Answer\textgreater\\ \\ Now you need to simplify the THOUGHT \\PROCESS as short as possible to only include \\the key information needed to solve the question.\\ And do not add additional information that is not \\included in the original THOUGHT PROCESS.\\ \\ SIMPLIFIED THOUGHT PROCESS:\end{tabular} \\ \bottomrule
    \end{tabular}



\end{table}

\begin{table}[h!]

    \centering

    \begin{tabular}{@{}l@{}}
    \toprule
    \textbf{\begin{tabular}[c]{@{}l@{}}Prompt For Compression with Specified \\ Compression Rate:\end{tabular}}                                                                                                                                                                                                                                                                                                                                                                                                                                                                 \\ \midrule
    \begin{tabular}[c]{@{}l@{}}You have a question now:\\ \\ QUESTION:\\ \textless{}Here is Instruction\textgreater\\ \\ THOUGHT PROCESS:\\ \textless{}Here is Original CoT\textgreater\\ \\ ANSWER:\\ \textless{}Here is Final Answer\textgreater\\ \\ Now you need to simplify the THOUGHT \\PROCESS to no more than\\ \textless{}Here is Word Numbers\textgreater words and retain the \\key information needed to solve the question.\\ And do not add additional information that is not \\included in the original THOUGHT PROCESS.\\ \\ SIMPLIFIED THOUGHT PROCESS:\end{tabular} \\ \bottomrule
    \end{tabular}



\end{table}

\begin{table}[hbt]

    \centering

    \begin{tabular}{@{}l@{}}
    \toprule
    \textbf{Prompt For Expansion:}                                                                                                                                                                                                                                                                                                                                                                                                                                                                                                                                                                                                                                                                                                                                                                                                                                                                                                                                                                                   \\ \midrule
    \begin{tabular}[c]{@{}l@{}}You have a question now:\\ \\ QUESTION:\\ \textless{}Here is Instruction\textgreater\\ \\ THOUGHT PROCESS:\\ \textless{}Here is Original CoT\textgreater\\ \\ ANSWER:\\ \textless{}Here is Final Answer\textgreater\\ \\ Now you need to expand the THOUGHT \\PROCESS according to the following \\STRATEGIES.\\ Do not remove anything from the original \\THOUGHT PROCESS.\\ \\ STRATEGIES:\\ 1. Think About The Word: pick important words \\in the QUESTION and interpret them.\\ 2. Read the Question Again: read the QUESTION \\repeatedly to reduce the interference of other texts \\on the chain of thought.\\ 3. Repeat State: add a small summary of the \\current state after a long chain of reasoning.\\ 4. Self-Verification: before getting the answer, \\add a self-verification process to judge whether \\the answer is reasonable based on some basic \\information.\\ 5. Make Equation: make equations whenever \\calculations are needed.\\ \\ EXPANDED THOUGHT PROCESS:\end{tabular} \\ \bottomrule
    \end{tabular}



\end{table}

%% file: fig_case_study_compressor.tex
\begin{figure*}[hbt]
    \begin{center}
    \includegraphics[width=\textwidth]{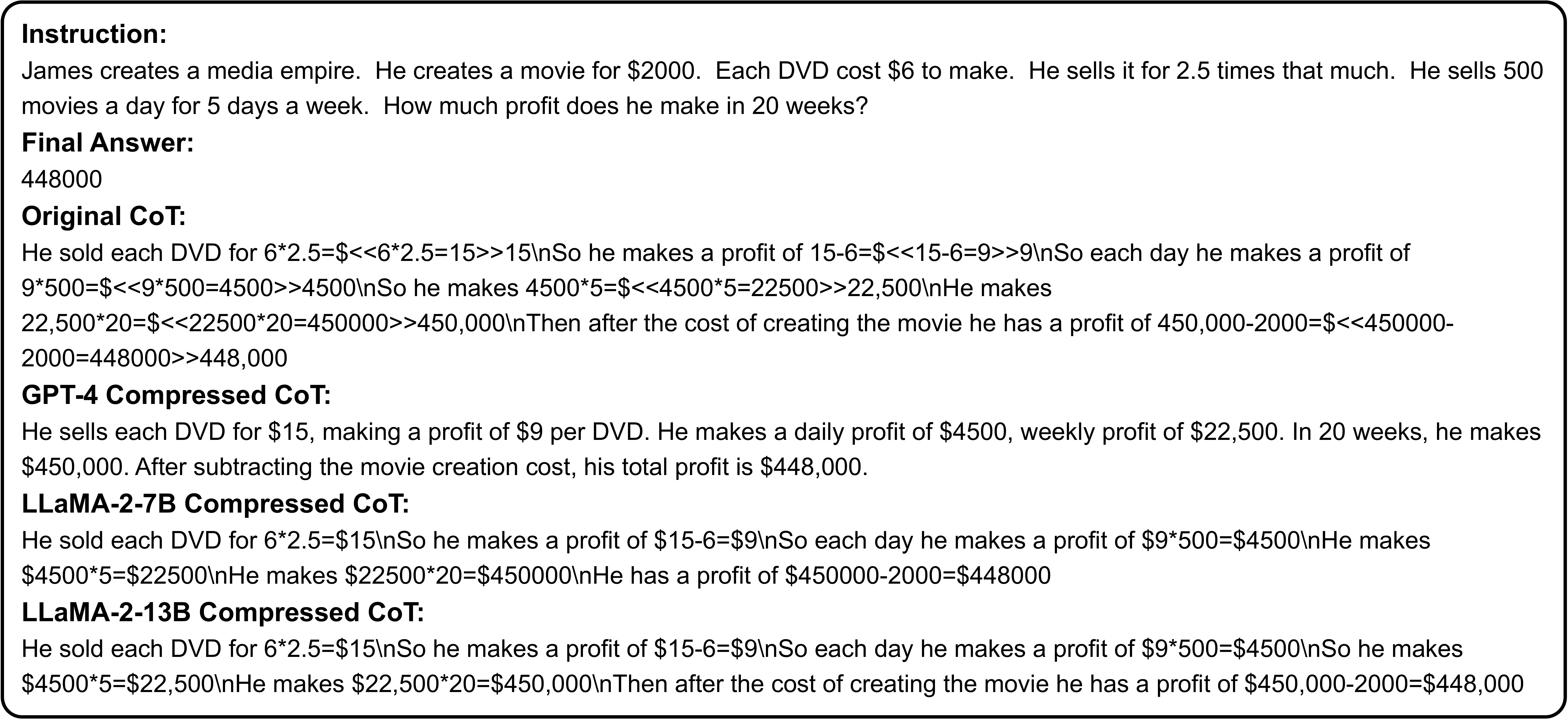}
    \end{center}
    \caption{Cases of Different Compressors}
\end{figure*}

\begin{figure*}[hbt]
    \begin{center}
    \includegraphics[width=\textwidth]{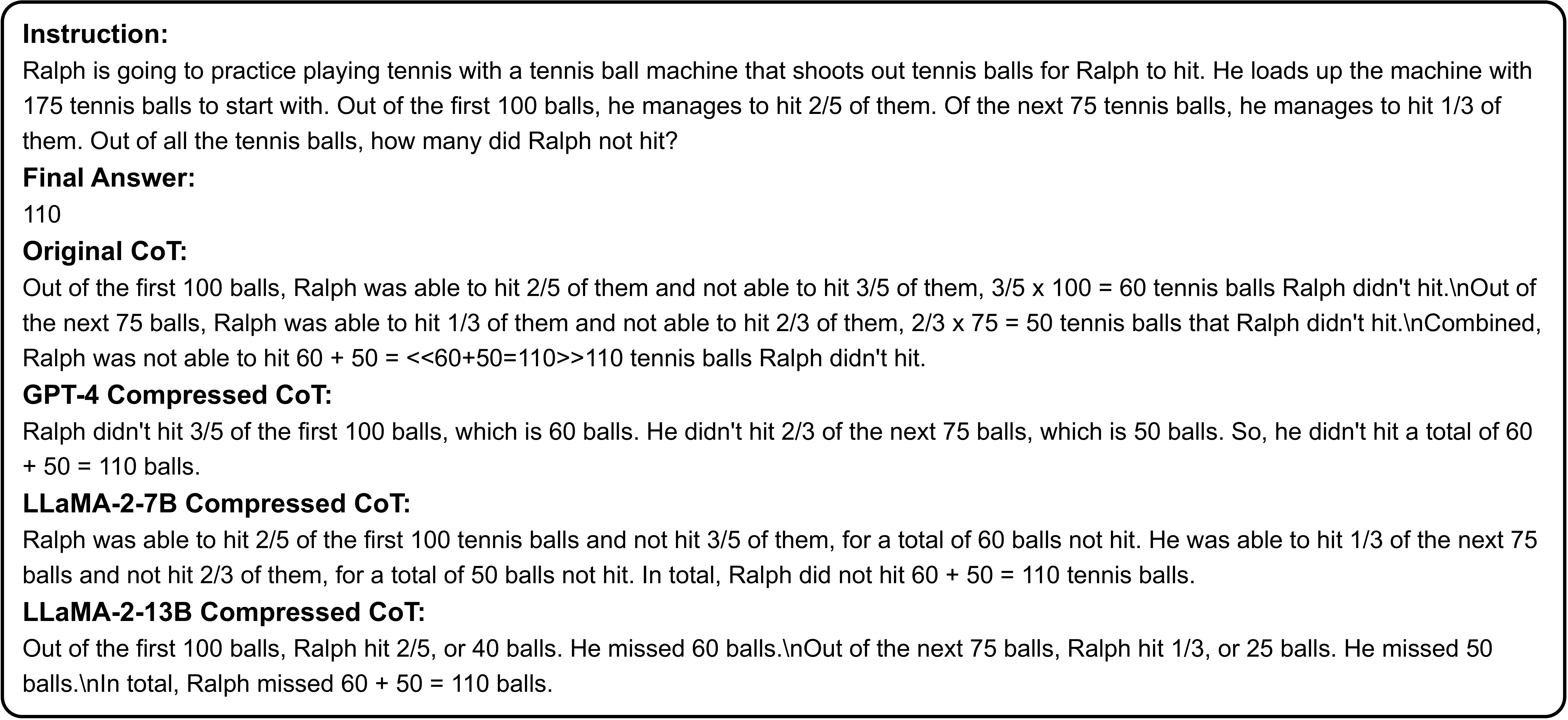}
    \end{center}
    \caption{Cases of Different Compressors}
\end{figure*}

%% file: fig_case_study_expansion.tex
\begin{figure*}[hbt]
    \begin{center}
    \includegraphics[width=\textwidth]{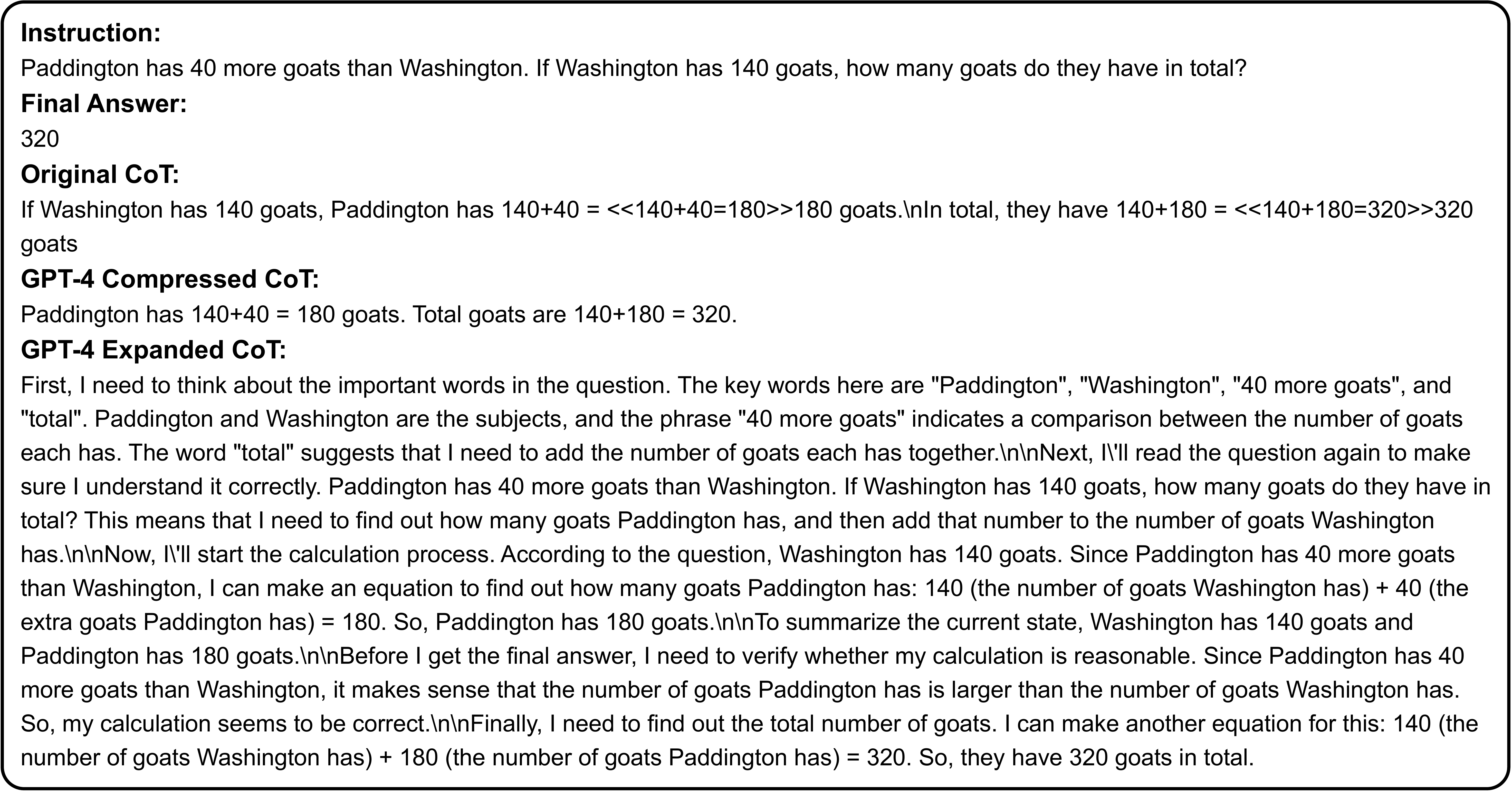}
    \end{center}
    \caption{Cases of CoT Expansion}
\end{figure*}

\begin{figure*}[hbt]
    \begin{center}
    \includegraphics[width=\textwidth]{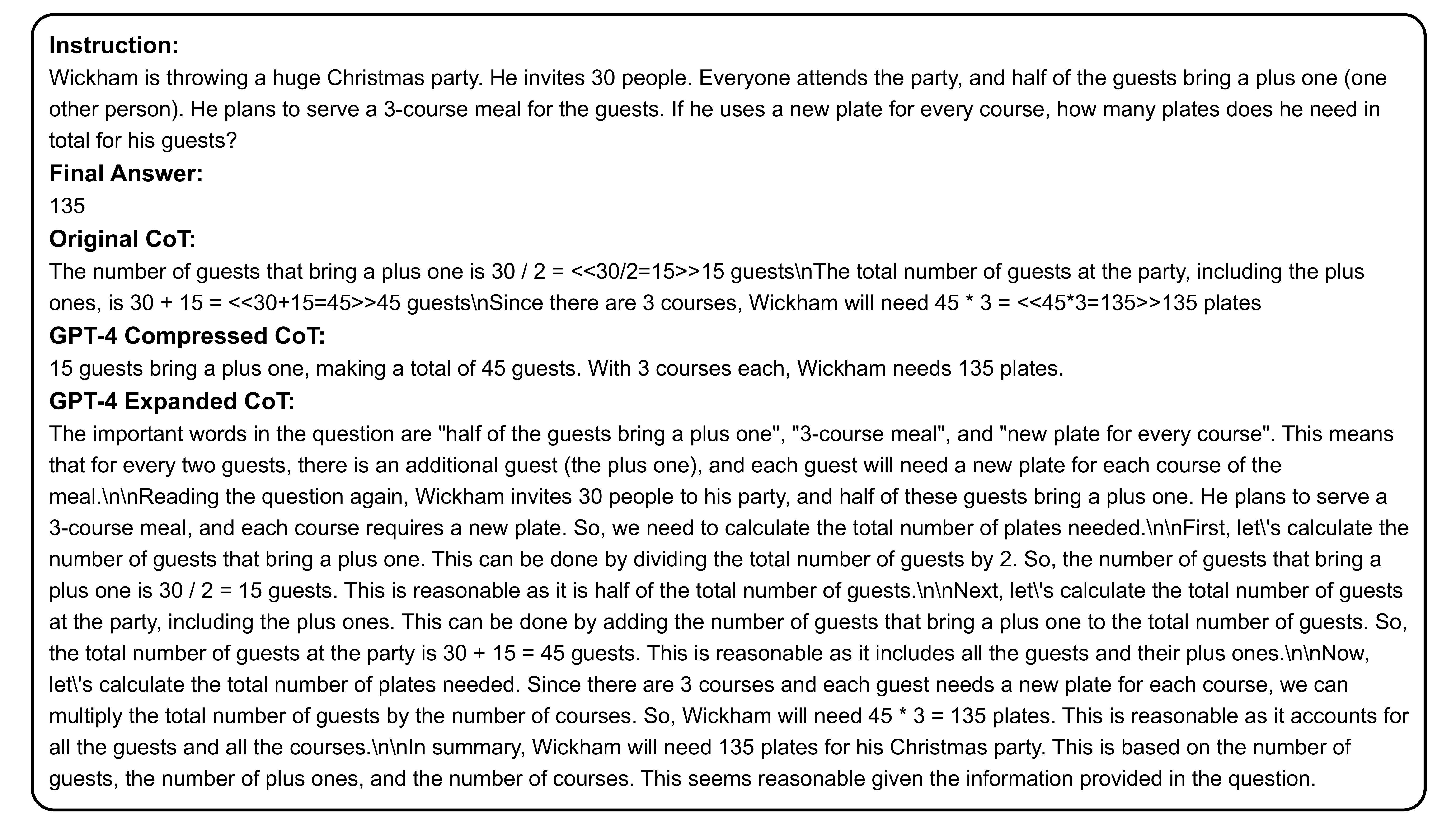}
    \end{center}
    \caption{Cases of CoT Expansion}
\end{figure*}

%% file: fig_case_study_rate.tex
\begin{figure*}[hbt]
    \begin{center}
    \includegraphics[width=\textwidth]{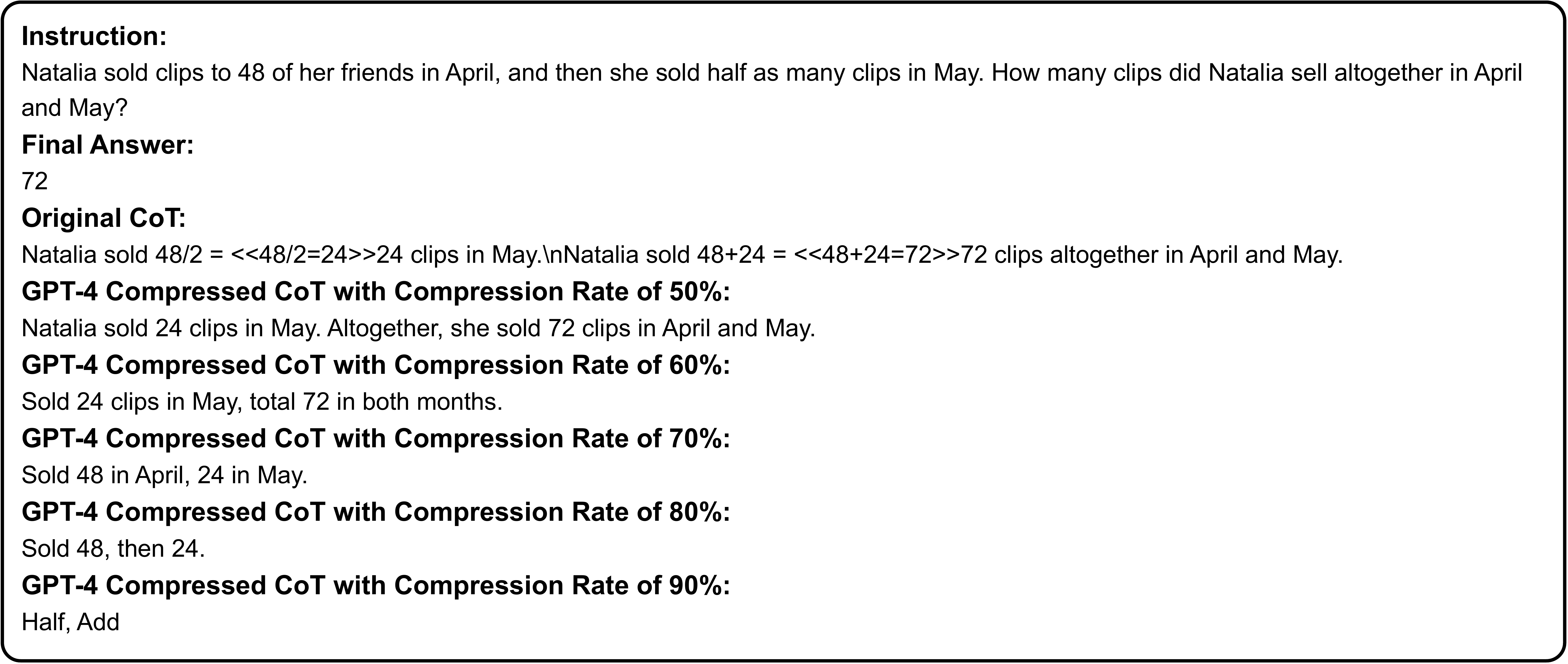}
    \end{center}
    \caption{Cases of Different Compression Rates}
\end{figure*}

\begin{figure*}[hbt]
    \begin{center}
    \includegraphics[width=\textwidth]{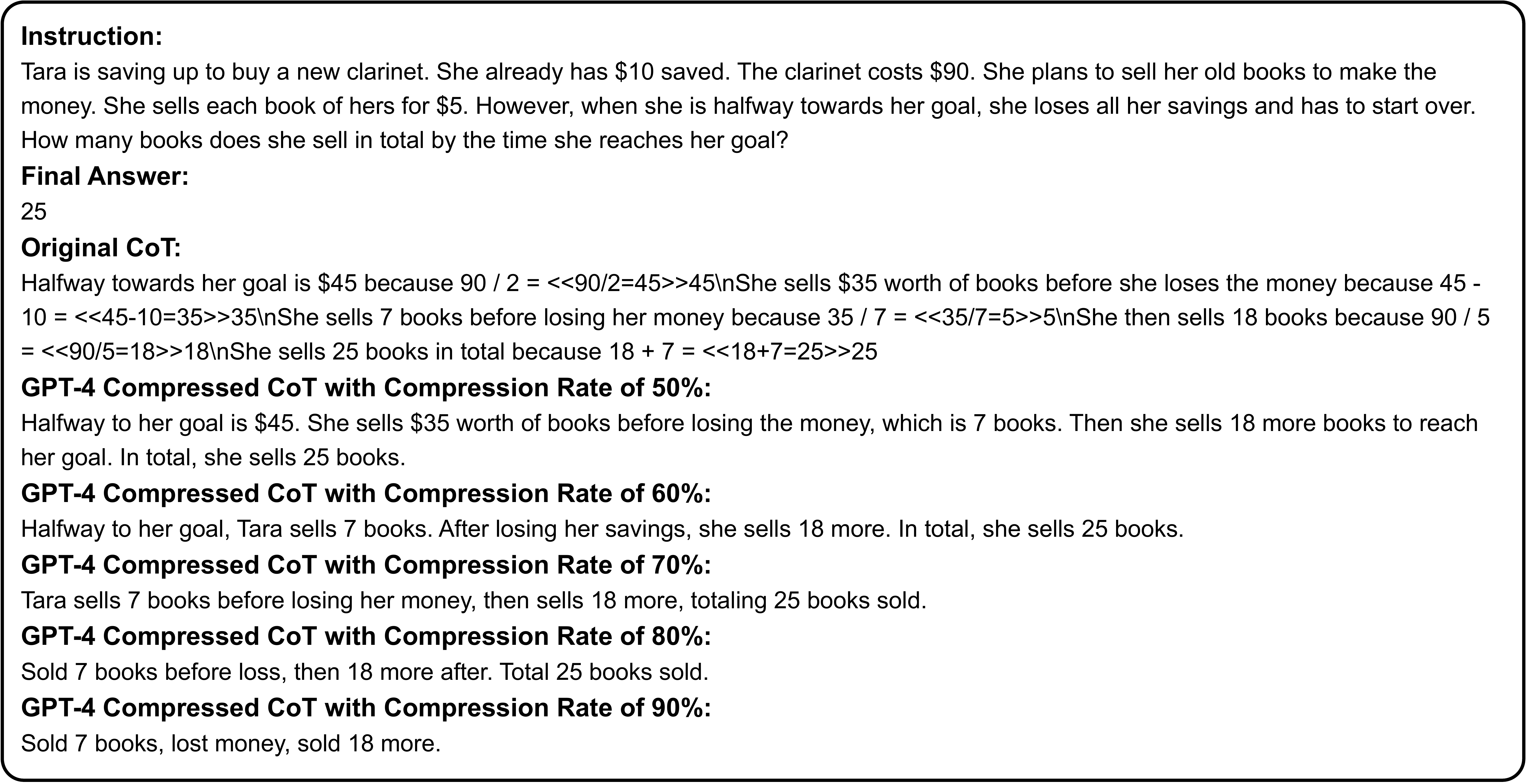}
    \end{center}
    \caption{Cases of Different Compression Rates}
\end{figure*}